\documentclass[runningheads]{llncs}
\pdfoutput=1
\usepackage{todonotes}
\usepackage{graphicx}
\usepackage[numbers]{natbib}
\newtheorem{RQ}{RQ}
\usepackage{sidecap}

\usepackage[caption=false]{subfig}
\usepackage{breqn}
\usepackage{caption}
\usepackage{pifont}
\usepackage{stfloats}
\usepackage{booktabs}

\begin{document}
\title{Finding Interpretable Concept Spaces in Node Embeddings using Knowledge Bases}
%
\titlerunning{Finding Interpretable Concept Spaces}
%
\author{Maximilian Idahl \and
Megha Khosla \and Avishek Anand}
\authorrunning{M. Idahl et al.}
%
\institute{L3S Research Center, Hannover \\
\email{\{idahl,khosla,anand\}@l3s.de}}
\maketitle              
\begin{abstract}
In this paper we propose and study the novel problem of explaining node embeddings by finding embedded human interpretable subspaces in already trained unsupervised node representation embeddings.
We use an external knowledge base that is organized as a taxonomy of human-understandable concepts over entities as a guide to identify subspaces in node embeddings learned from an entity graph derived from Wikipedia.
We propose a method that given a concept finds a  linear transformation to a subspace where the structure of the concept is retained.
Our initial experiments show that we obtain low error in finding fine-grained concepts.


\end{abstract}

\section{Introduction}
\label{sec:intro}

Representations of nodes in a graph or node embeddings have proven useful in many applications such as question answering~\cite{bordes2014question}, dialog systems~\cite{ma2015knowledge}, recommender~\cite{palumbo2018empirical} systems and knowledge-base completion~\cite{meilicke2018fine}. 
The core idea behind \textit{node representation learning} (NRL)~\cite{perozzi2014deepwalk,grover2016node2vec,khosla2019node} approaches is to distill the high-dimensional discrete representation of nodes into a dense vector embedding using dimensionality reduction methods, which optionally not only incorporate the graph structure, but also features attached to nodes. 
These representations can be seen as features extracted from only the topology or from both the topology and the available node attributes. 
The dense representations thereby learnt form a latent feature space where the basis or dimensions are non-interpretable.


Consequently, in spite of their success, there is a lack of an understanding of what the latent dimensions encode in terms of existing human knowledge. This is problematic for downstream tasks requiring interpretability, since using such embeddings results in the input already being non-interpretable.
For aiding interpretability and utility of these embeddings in downstream application scenarios we initiate an inquiry into presence of \emph{interpretable} or human understandable subspaces in the learnt feature representation space of these graph embeddings.
We ask the fundamental question: \emph{What do node embeddings encode in terms of human world knowledge?} 
Recent works in interpretability for learning on structured data either focus on generating interpretable embeddings or explaining the predictions made by a classifier to which embeddings form the input~\cite{ying2019gnn}. But none of these methods provide insights into the embedding itself, a problem which we propose and study in this work.

We take an alternate view on interpretability of node embeddings in that we want to find sub-spaces in the embedding space corresponding to human-understandable concepts.
Our main contribution is in finding interpretable sub-spaces in the latent feature representation space and thus characterizing the behavior of node representations when projected into these interpretable spaces.
This has two distinct advantages -- first we do not compromise on the effectiveness of these embeddings as we post-hoc analyze the presence of interpretable spaces in the already learned representation space.
Secondly, we ground the interpretable space to existing world knowledge in the form of knowledge bases.

To this extent, in this work, we use external knowledge bases (KB) to learn conceptual spaces for corresponding characteristics that can be attributed to a given node. 
In particular, we assume that we have an input graph of labelled or named nodes. As a use case we focus on a hyperlink graph of named entities. 
We observe that KBs like YAGO~\cite{hoffart2011yago2} encode human understandable concepts organized in a taxonomy which can be used as the source of world knowledge assuming that the nodes/entities in the input graph are also present in the taxonomy.
In principle one can use any input graph and KB as long as the input graph node names are grounded in the KB. 
Having extracted the possible concepts from the taxonomy, we then propose methods to explain a node embedding in terms of the applicability of various concepts. For example, a node named \emph{Albert Einstein} could be explained by concepts like \emph{Theoretical physicists}, \emph{Scientists} etc.  

We  propose two simple algorithms, SAS and CSD, to explain node embeddings in terms of concepts and provide promising first results for pre-trained embeddings corresponding to two unsupervised random walk based node embedding methods, namely, DeepWalk~\cite{perozzi2014deepwalk} and LINE~\cite{tang2015line}. 
We show that our second approach CSD that projects a node embedding to a common learnt concept space distinguishes the applicable and non applicable concepts better than our first approach which operates in the original embedding space. 




\section{Related Work}
Supervised learning approaches are either \emph{interpretable by design}~\cite{caruana2015intelligibletrees,rulesletham2015interpretable,captioningxu2015showattention} or explanations can be generated in a post-hoc manner after the model is trained~\cite{ribeiro2016should,influencefunctionskoh2017understanding,montavon2017explaining}.
Post-hoc methods for interpretability either operate introspectively (full access to the model parameters)~\cite{influencefunctionskoh2017understanding,montavon2017explaining} or are model agnostic~\cite{ribeiro2016should}. 
We operate in the model introspective interpretable regime where we assume full access to the model parameters.
For other notions of interpretability and a more comprehensive description of the approaches we point the readers to~\cite{guidotti2018survey}.

Methods focussing on building interpretable representations include MEmbER~\cite{jameel2017member} which learns entity embeddings using max-margin constraints to encode the desideratum that (salient) properties of entities should have a simple geometric representation in the entity embedding. Jameel and Schockaert~\cite{jameel2016entity} propose a method which learns a vector-space embedding of entities from Wikipedia and constrains this embedding such that entities
of  the  same  semantic  type  are  located  in  some  lower-dimensional
subspace. Minervini et al.~\cite{minervini2017regularizing} leverage equivalence and
inversion axioms during the learning of knowledge graph embeddings, by imposing a set of model dependent soft constraints on the predicate embeddings. Post-hoc methods include GNN-Explainer~\cite{ying2019gnn} which provides interpretations for GNN predictions on link prediction, node classification and graph classification tasks. The interpretations are tied to specific tasks. We, on the other hand, propose to understand the node representations directly in terms of user provided conceptual categories. 

Unlike the above works we focus on explaining the node vector representation itself which might have been obtained using an arbitrary embedding method.

\section{Preliminaries}
\label{sec:prel}
 In this section we give a brief overview of YAGO and node embedding methods used in this work.
\subsection{Knowledge Graphs}
\label{sec:KB}
As a source of \emph{concepts} or human understandable world knowledge we use the YAGO~\cite{hoffart2011yago2} knowledge base (KB), which was automatically constructed from Wikipedia. Typically, each article in Wikipedia becomes an entity in the knowledge base (e.g., since Albert Einstein has an article in Wikipedia, Albert Einstein is an entity in YAGO). Each entity is organized into a taxonomy of classes. In addition, every entity is an instance of one or multiple classes and every class (except the root class) is a subclass of one or multiple classes. therefore yielding a hierarchy of classes — the \emph{YAGO taxonomy}. 

Each class name is of the form \texttt{<wordnet\_XXX\_YYY>} or \texttt{<wikicat\_XXX\_YYY>} , where XXX is the name of the concept (e.g., singer), and YYY is the WordNet 3.0 synset id of the concept (e.g., 110599806). For example, the class of singers is \texttt{<wordnet\_singer\_110599806>}. 
Additionally, each class is connected to its more general class by the \texttt{rdfs:subclassOf} relationship. 

Not all Wikipedia categories correspond to classes in YAGO.
The lowest layer of the taxonomy is the layer of instances. 
Instances comprise individual entities such as rivers, people, or movies. For example, the lowest layer contains \texttt{<Elvis\_Presley>}. 
Each instance is connected to one or multiple classes of the higher layers by the relationship rdf:type. 
For example, for entity Albert\_Einstein we have: 
$$
\texttt{<Albert\_Einstein>} \,\,\,\,\,\, \texttt{rdf:type} \,\,\,\,\,\, \texttt{<wikicat\_Nuclear\_physicist>}.
$$
One can therefore walk from the instance up to its class by \texttt{rdf:type}, and then further up by \texttt{rdfs:subclassOf}. In Section \ref{sec:approach} we will provide details about how the concepts derived from the taxonomy are used as explanations for node embeddings.

\subsection{Node Embeddings} 
\label{sec:node_emb}
Node representations or node embeddings can be understood as the set of features extracted from the graph topology and (if given) node attributes. The present set of techniques for node representation learning generally fall into one of these categories : (1) random walk based ~\cite{perozzi2014deepwalk,grover2016node2vec,tang2015line,khosla2019node}, (2) matrix factorization based ~\cite{cao2015grarep,ou2016asymmetric} or (3) deep learning or Graph Neural Network (GNN) based ~\cite{kipf2016semi, hamilton2017inductive}. In this section we describe briefly the two random walk based approaches which we employ in this work. In future we will investigate our methods using a general set of unsupervised and semi-supervised embedding approaches. 

The basic idea behind random walk based embedding techniques is to transform the graph into a collection of node sequences, in which, the occurrence frequency of a node-context pair measures the structural distance between them. \emph{DeepWalk}~\cite{perozzi2014deepwalk} was the first method to exploit random walk techniques to build sentence like structures from graphs to train  a \emph{SkipGram} model ~\cite{mikolov2013distributed}. It employs truncated random walks to create vertex sequences, which are later used in a word2vec fashion to learn vertex embeddings given its context. For a graph $G$, it samples  uniformly  a  random  vertex $v$
as  the  root  of  the  random walk $W_v$.  A walk samples uniformly from the neighbors of the last vertex visited until the maximum length $t$ is
reached. For each $v_i \in W_v$ and for each $u_k \in W[j-c:j+c]$ (c is the window size), $(v_j,u_k)$ forms a vertex-context training pair (similar to word -context pair in word embeddings). The objective is then to maximize the probability of observing $u_k$ given the representation of $v_j$. LINE~\cite{tang2015line} optimizes first order proximity ( i.e. embeds nodes sharing a link closer) and second order proximities (embeds nodes closer if they have similar neighborhoods) using an SGNS (Skip-gram with negative sampling) objective function~\cite{mikolov2013efficient}. Similar to DeepWalk, it can be understood as sampling random walks of length 1 and uses vertices sharing an edge as training pairs.





\section{Research Questions and our Approach}
\label{sec:approach}
We propose a general approach for post-hoc interpretability of node representation learned by an unsupervised or semi-supervised method. We bring in a completely new perspective of interpretability of extracted features of nodes by using external knowledge to determine the concepts that a given representation encodes. More precisely, we use Wikipedia entity graph, $G=(V,E)$, as the input graph, where the nodes are Wikipedia pages and the edges correspond to the hyperlinks between them. We employ DeepWalk and LINE to generate embeddings for all $v\in V$. We ignore the edge direction to learn the embeddings. We also recall that the present topic of this work is to define and validate interpretability on node embeddings and the choice of embeddings methods is therefore arbitrary.  Let $\Phi_v$ represent the embedding vector corresponding to $v$.
We ponder over the following question:
\begin{RQ}What concepts do these embeddings encode?  
\end{RQ}
As the embeddings are usually generated only considering the structure of the graph or/and node attributes, an embedding vector $\Phi_v$ encodes the concepts which it shares with its neighborhood (neighborhood here depends on the employed embedding method). Consider, for example, an entity \emph{Barack Obama}, which could be understood as sharing characteristics with other \emph{Presidents} and \emph{Nobel Prize winners}. \emph{Presidents} and \emph{Nobel Prize winners} here are the human understandable world knowledge or concepts. Rather than characterizing nodes in terms of their neighbors, we in this work use these implicit human understandable concepts to characterize an embedding vector. In particular, for a given embedding vector $\Phi_v$ and a concept $c$, we assign a score $\mathcal{S}(\Phi_v, c)\in \mathcal{R}$ which quantifies the characteristic $c$ of the embedding $\Phi_v$. Roughly speaking, the score measures the amount of the characteristic that an embedding vector possesses. 

The challenge here is that often only the graph structure or sometimes the node attributes are also available but there are no explicit concepts provided. We therefore ask the following question:
\begin{RQ}
How can explicit concepts be constructed given an input graph with named vertices?
\end{RQ}

In order to generate possible concepts related to an entity, we propose the use of external knowledge base like YAGO\emph{} (see also Section~\ref{sec:KB}), which provides a hierarchy of concepts related to any given node, say $v$ in the graph. These concepts form the characteristics of $v$.
The user can then query the encoding of possible concepts in the trained node embedding. For example, a user may ask how much the embedding vector corresponding to \emph{Barack Obama} encodes \emph{American Presidents} and \emph{Scientists}. One might assume that the Obama's embedding vector should not have anything to do with the concept \emph{Scientists}, which might not be true as the underlying graph might put Obama in close proximity with other Nobel Prize winners who are also Scientists. Having defined or collected concepts from external knowledge bases, the next natural question is:

\begin{RQ}
For a given embedding vector, $\Phi_v$ and a concept $c$, how can we score the applicability of $c$ to $\Phi_v$ ? 
\end{RQ}
To quantify the applicability of concept corresponding to an embedding or to explain an embedding in terms of the applicable and not applicable concepts, we propose two algorithms: Simple Aggregation Strategy (SAS) and Concept Space Discovery (CSD).

\subsection{Simple Aggregation Strategy}
The first approach uses a simple aggregation strategy to build concept representations from the representations of the nodes (from the training set) to which the concept is applicable (test nodes are held out). In particular, we first compute a vector representing the given concept by taking the element-wise mean of all the  embedding vectors corresponding to nodes to which the concept applies, excluding the query nodes. This vector defines the \textit{concept center}. To score a query node, we compute the L2 distance between its embedding vector and the concept center.

\begin{figure}[!ht]
    \centering
    \includegraphics[width=0.8\textwidth]{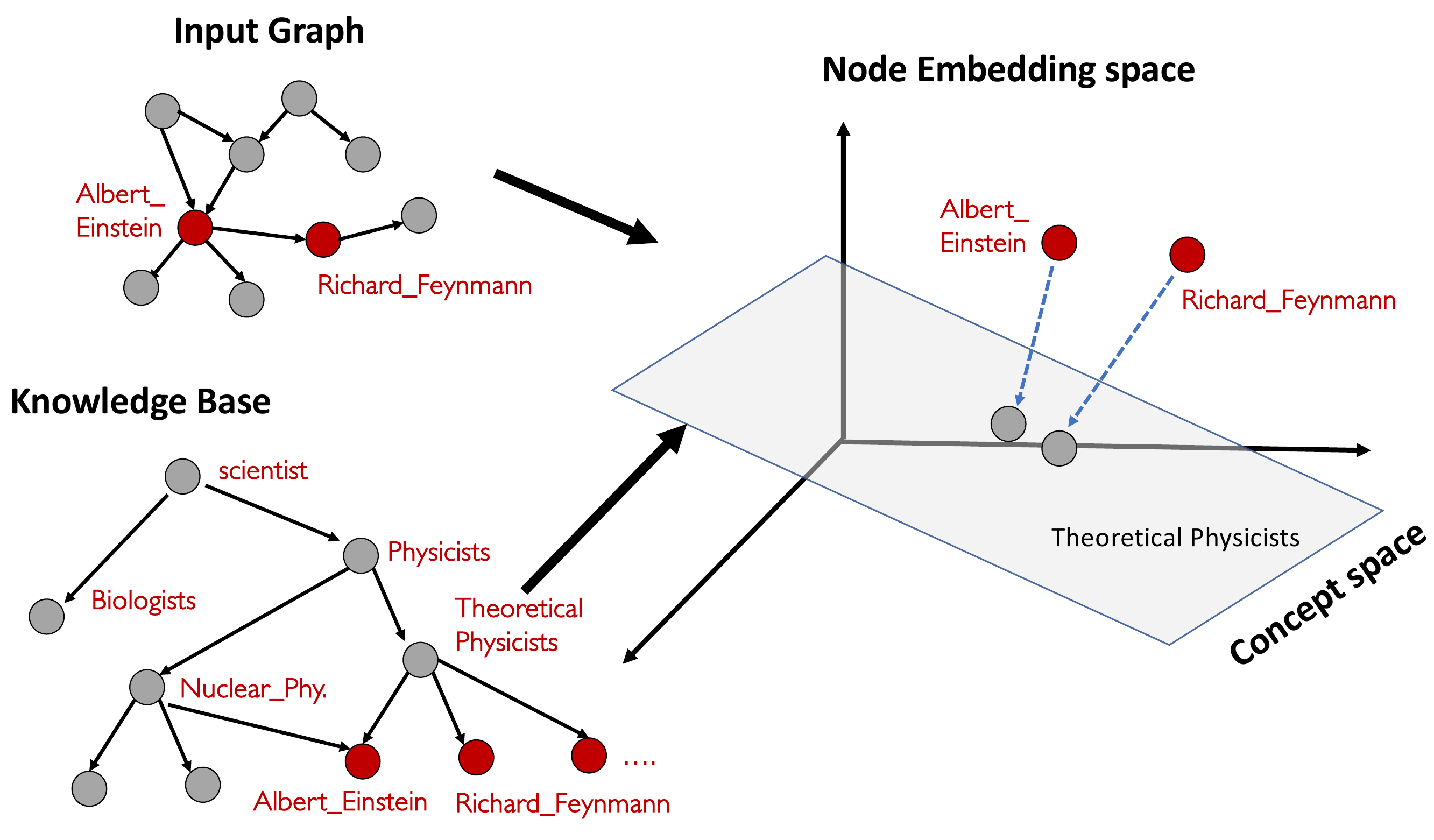}
    \caption{Extracting Concept Spaces}
    \label{fig:concept_spaces}
    \vspace{-5mm}
\end{figure}
\subsection{Concept Space Discovery (CSD)} 
\label{sec:CSD}
\begin{figure}[!ht]
    \centering
    \includegraphics[height=0.3\textwidth]{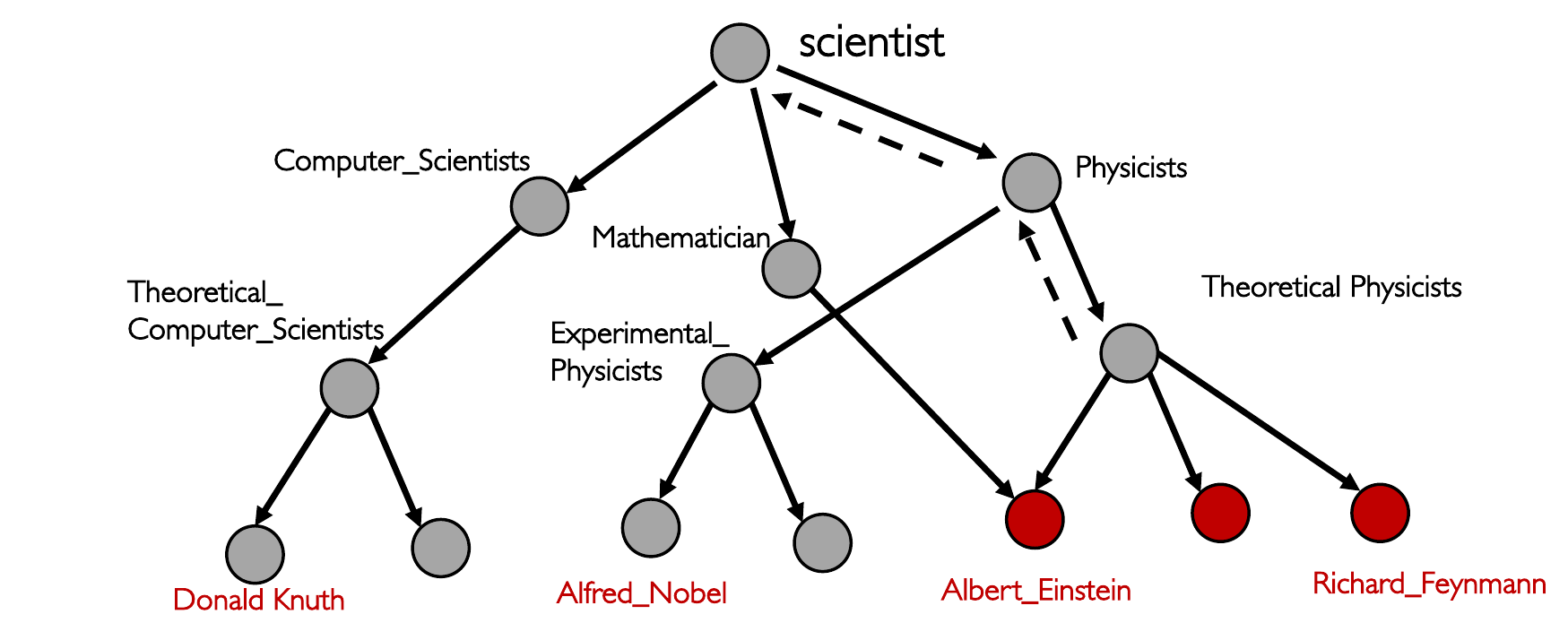}
    \caption{Margin for Triplet loss is determined by the similarity in the taxonomy graph. The margin between Albert Einstein and Donald Knuth is 2, where as the margin between Albert Einstein and Alfred Nobel is 1.}
    \label{fig:triplet_loss_vis}
    \vspace{-5mm}
\end{figure}

The second algorithm is more involved and explicit, in the sense that for each concept $c$ it learns a linear transformation, which is used to project the node vectors into a more restricted space for $c$, that we call \textit{concept space}. 
The original embedding vectors are projected into this new space to extract their effective representations which best encode the given concept (refer Figure~\ref{fig:concept_spaces}). 
We learn the parameters for this transformation on triplets of entities, using triplet loss. Let $a$ be the entity node (also called anchor node)  which is a direct descendant of concept $c$, $p$ be some sibling of $a$ in the taxonomy and $n$ be the negative example , i.e., an entity  which is not a sibling of $a$ in the taxonomy. For any node $v$, let $\Phi_v$ represent the corresponding embedding vector. The triplet loss $ \mathcal{L}(a,p,n)$ is then defined as follows.
\begin{equation}
    \mathcal{L}(a,p,n) = max\{d(\Phi_a,\Phi_p) - d(\Phi_a,\Phi_n) + m, 0\}
\end{equation}
where $d(\Phi_x,\Phi_y) = ||x - y||_2$ and $m$ is a margin specific to the negative entity in a triplet. We set this margin to be the distance from the target concept to the lowest common ancestor concept shared by the positive and the negative entity, i.e. negative entities that are conceptually close to the positive entity have lower margins and ones that are conceptually far away have higher margins. We refer to negative entities with low margins as soft negatives and to negative entities with high margins as hard negatives.
An illustrative example for computing margin is provided in Figure~\ref{fig:triplet_loss_vis}.


\textbf{Score Computation.} The scoring of how much a concept applies to a query entity is analogous to the first approach, but of course operates in the \textit{concept space}. 
That is, for a given concept $c$ and the positive entities (the training set) corresponding to the concept, we first compute their projections into common concept space and then compute the mean of the resulting projected vectors to represent the concept. Again for a given query node, we first compute its projection into the concept space and the final score is then given  by the L2 distance between the concept vector and the query projection. Lower the score, better is the concept encoded by the query node.
Note that both loss function and scoring make use of the same distance metric, the L2 distance.

\section{Experiments}
\subsection{Data Acquistion}
We conduct our experiments on the Wikipedia entity graph, where the nodes are Wikipedia pages and the edges correspond to the hyperlinks between them. In addition, we use the type hierarchy of YAGO as the KB and consider all leaves under a concept node as belonging to the concept, as described in Section~\ref{sec:KB}.






\subsection{Methodology}
Given a query entity $q$ and a start concept $c_{start}$ we learn concept spaces for $c_{start}$ and its sibling concepts in the taxonomy. Note that we limit the number of concepts due to computation (Some concepts have a large number of siblings). For each selected concept, we a learn a concept representation as described in Section ~\ref{sec:approach}. Below we give more details about the training employed in our second approach CSD.

For CSD where we use triplet loss function to learn the concept space we choose positive and negative examples as follows. For each concept $c$, the set of positive entities (examples) consists of all entities contained in $c$. Next, we rank all ancestor concepts of $c$ by the margin, which is the distance of the concept to $c$. Following Figure \ref{fig:triplet_loss_vis}, if $c$ is \textit{Theoretical Physicists}, then entities which belong to the concept \textit{Physicists} are negative entities with a margin of 1, entities belonging to the concept \textit{Scientists} are negative entities with a margin of 2, and so on. Note that an entity is always assigned the lowest possible margin. In this example, all physicists get assigned a margin of 1 and only all scientists that are not physicists get assigned a margin of 2. We also exclude the query entity $q$ from the sampling process.
We split the sets of positive and negative entities into a training and a validation set, taking 20\% of the entities for the validation set. 

In order to generate a triplet, we select a positive entity uniformly from the set of positive samples. An anchor entity is selected in the same way, with respect to the anchor not being the same entity as the chosen positive one. Next, we select a margin $m$ uniformly from the available margins in the set of negative entities. Then, we select a negative entity uniformly from the negative samples corresponding to margin $m$. To train one concept space, we sample a total of ten thousand triplets. We then train the linear transformation using Stochastic Gradient Descent with Momentum for 100 epochs, with a mini batch size of 16 and a leaning rate of 0.001. We stop the training early if the validation loss does not improve over 5 epochs. After training, we score the query entity as described in Section~\ref{sec:approach} corresponding to our two approaches.

\begin{SCfigure}
  \caption{Mean validation losses for training concept space projections for concepts of different hierarchy levels. Level 1 includes concepts high up in the hierarchy, namely \textit{person}, \textit{organization} and \textit{country}. The second level includes \textit{scientist}, \textit{educational institution} and \textit{countries in Europe}. Level 3 then covers the more fine-grained concepts \textit{theoretical physicist}, \textit{university or college in Germany} and \textit{states of Germany}.}
  \includegraphics[width=0.5\textwidth]{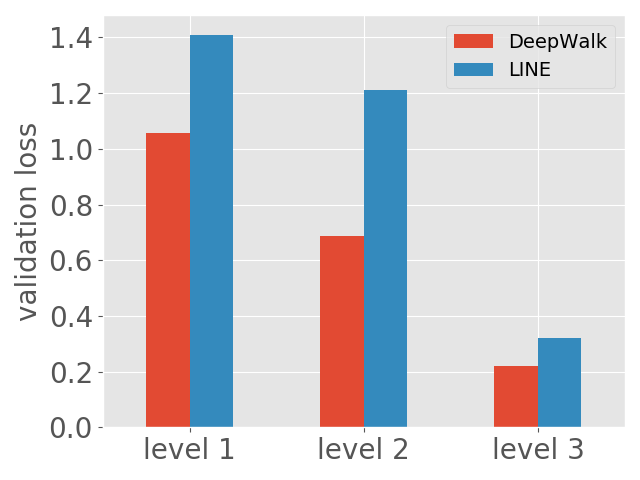}
  \label{fig:val}
\end{SCfigure}

\begin{figure*}[ht!]
\centering

    \subfloat[SAS DeepWalk]{\includegraphics[width=0.5\textwidth]{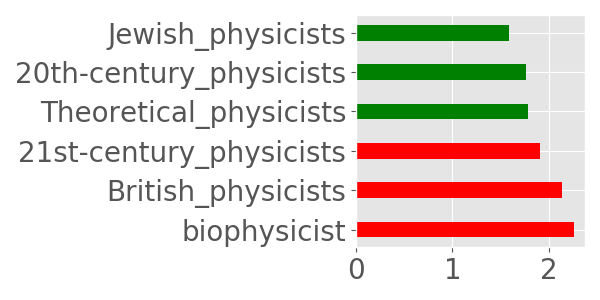}}
    \subfloat[SAS LINE]{\includegraphics[width=0.5\textwidth]{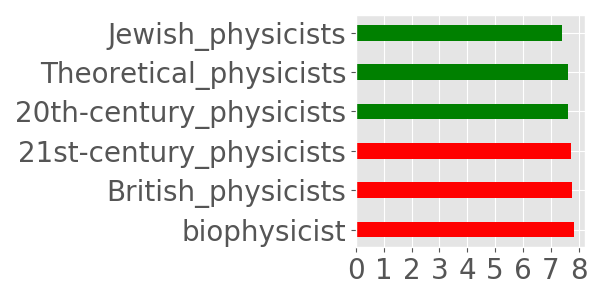}}
    \hfill
    \subfloat[CSD DeepWalk]{\includegraphics[width=0.5\textwidth]{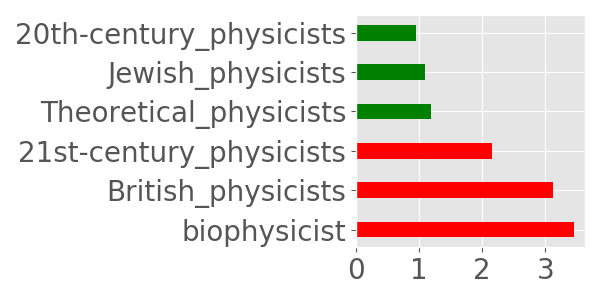}}
    \subfloat[CSD LINE]{\includegraphics[width=0.5\textwidth]{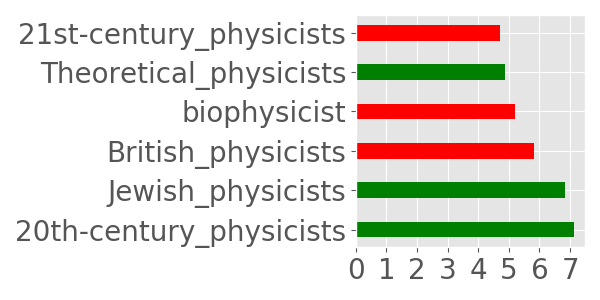}}
     
\caption{Concept Ranking for Albert Einstein}
\label{fig:ae}
\end{figure*} 

\begin{figure*}[ht]
\centering

    \subfloat[SAS DeepWalk]{\includegraphics[width=0.5\textwidth]{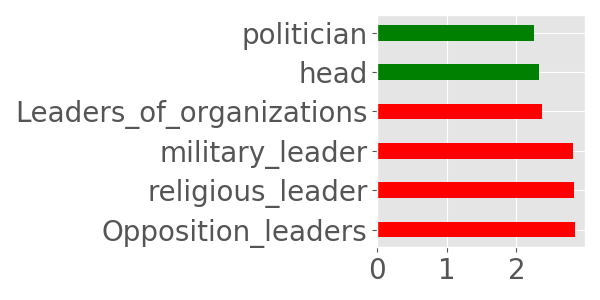}}
    \subfloat[SAS LINE]{\includegraphics[width=0.5\textwidth]{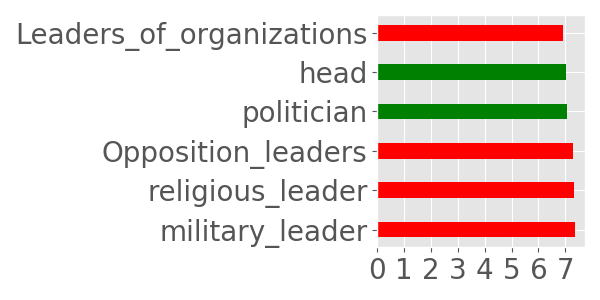}}
    \hfill
    \subfloat[CSD DeepWalk]{\includegraphics[width=0.5\textwidth]{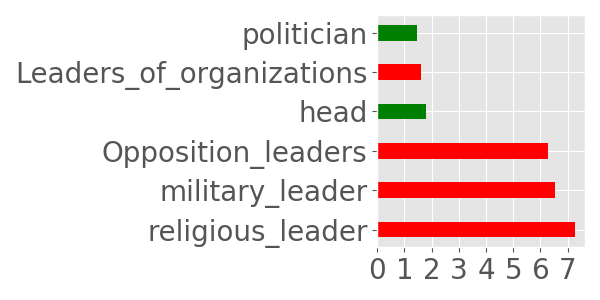}}
    \subfloat[CSD LINE]{\includegraphics[width=0.5\textwidth]{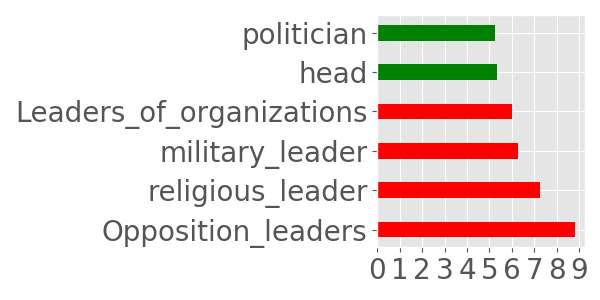}}
     
\caption{Concept Ranking for Donald Trump}
\label{fig:dt}
\end{figure*}

\section{Results}
\label{sec:results}
In Figure~\ref{fig:val} we show the errors corresponding to each concept level for different node embedding approaches (DeepWalk and LINE). Concepts at a higher level, as expected, exhibit higher error but the error reduces to a small value for more specific concepts. It is interesting to observe that it is easier to find interpretable concept spaces in DeepWalk as opposed to LINE. In this regard DeepWalk can be in some sense regarded as more interpretable than LINE.


Figure \ref{fig:ae} and Figure \ref{fig:dt} show the scores of different concepts for the query entities \textit{Albert Einstein} and \textit{Donald Trump}, respectively. We recall that lower the score $\mathcal{S}(\Phi_v,c)$, more is the applicability of $c$ towards the embedding vector $\Phi_v$ or the entity $v$.
Concepts under which the query entity is listed in YAGO are shown in green, concepts under which it is not listed in red. 

We note that for the query entity \textit{Albert Einstein}, scoring concepts in both of the original embedding spaces (Figure \ref{fig:ae} a,b) yields a correct ranking of the concepts. Yet, there is not much difference between the scores of concepts which apply to the query entity and the scores of non-applicable concepts. This is more prominent the case for the embeddings generated by LINE, where differences in the scores are barely noticeable. 

We observe a similar behaviour with our second query entity \textit{Donald Trump}. An interesting observation here is that the best ranked concept in Figure \ref{fig:dt} b, \textit{Leaders of organizations} which is not listed as applicable concept in the taxonomy, in fact applies to the query entity \textit{Donald Trump}. This is another finding, in the sense that the embeddings encode knowledge not present in YAGO. Using concept spaces to score the query entity increases the differences between scores. This seems to work well for both query entities when using the embeddings generated by DeepWalk. The concept spaces deliver scores where it is much clearer whether a concept applies to the query entity or not, as there is a large gap between applicable ones and non-applicable ones.

\section{Conclusions and Future Work}
\label{sec:conclusions}

In this work we proposed a method to find interpretable concept spaces for graph embeddings. 
We hypothesize that latent feature spaces that embed named vertices are not interpretable themselves but contain subspaces that do contain human understandable concepts.
We propose an algorithm that tries to find subspaces in the feature representation space by exploiting similarity of entities in the KB using triplet loss. 
We anecdotally show the effectiveness of our approach on a small subset of concepts chosen from the KB.

As future work there are plenty of avenues to investigate in detail. 
First, we would want to improve our evaluation procedure to quantitatively establish the effectiveness of our concept space discovery approach. This would require us to not only experiment with a large set of concepts but increase our coverage to multiple unsupervised and semi-supervised node representation learning methods. 
Secondly, we would want to find out that if there are non-linear sub spaces that encode coarse-granularity concepts like scientists, politicians etc. Currently, we see room for improvement in finding subspaces for coarser granularity topics due to choice of linear subspaces.

 \bibliographystyle{splncs04}
\bibliography{references}

\end{document}